\documentclass[10pt,twocolumn,letterpaper]{article}
\usepackage[pagenumbers]{wacv}
\usepackage{times}
\usepackage{epsfig}

\usepackage{graphicx}
\usepackage{amsmath}
\usepackage{amssymb}
\usepackage{booktabs}
\usepackage{multirow}
\usepackage{wrapfig}
\usepackage[pagebackref,breaklinks,colorlinks]{hyperref}

\usepackage[capitalize]{cleveref}
\crefname{section}{Sec.}{Secs.}
\Crefname{section}{Section}{Sections}
\Crefname{table}{Table}{Tables}
\crefname{table}{Tab.}{Tabs.}
\usepackage[table,xcdraw]{xcolor}
\newcommand{\sti}{$\textit{STI tokens}$}

\begin{document}

\title{Modelling Spatio-Temporal Interactions For Compositional Action Recognition}

\author{Ramanathan Rajendiran \\ \small Centre for Frontier AI Research, \\ \small Agency for Science, Technology and \\ \small Research (A*STAR), Singapore
\and
Debaditya Roy \\ \small Institute of High Performance Computing, \\ \small Agency for Science, Technology and \\ \small Research (A*STAR), Singapore
\and Basura Fernando \\ \small Centre for Frontier AI Research, \\ \small Agency for Science, Technology and \\ \small Research (A*STAR), Singapore}

\maketitle

\begin{abstract}
Humans have the natural ability to recognize actions even if the objects involved in the action or the background are changed.
Humans can abstract away the action from the appearance of the objects which is referred to as compositionality of actions. 
We focus on this compositional aspect of action recognition to impart human-like generalization abilities to video action-recognition models. 
First, we propose an interaction model that captures both fine-grained and long-range interactions between hands and objects. 
Frame-wise hand-object interactions capture fine-grained movements, while long-range interactions capture broader context and disambiguate actions across time. 
Second, in order to provide additional contextual cues to differentiate similar actions, we infuse the interaction tokens with global motion information from video tokens. 
The final global motion refined interaction tokens are used for compositional action recognition. 
We show the effectiveness of our interaction-centric approach on the compositional Something-Else dataset where we obtain a new state-of-the-art result outperforming recent object-centric methods by a significant margin.
\end{abstract}

\section{Introduction}
\label{sec:intro}
\begin{figure}
    \centering
    \includegraphics[width=\linewidth]{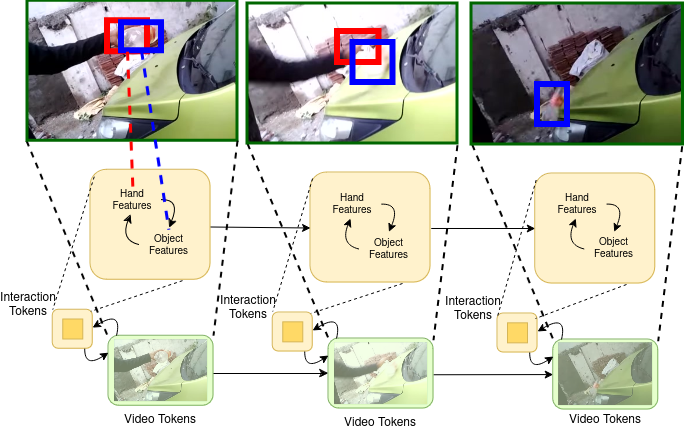}
    \caption{In order to recognize the compositional action \textit{letting something roll down a slanted surface}, one needs to pay attention to the local interactions involving the spatial positions and appearance features of the hand and the active object(s) in the scene. The local interaction tokens (yellow blocks) encode the changes in the coordinates and appearance features extracted from the hand (red) and the active object (blue). Furthermore, for the action \textit{rolling down}, the context of the \textit{slanted surface} of the car differentiates the motion from other similar actions such as \textit{moving something down} or \textit{lifting up one end of something, then letting it drop down}. Hence the local interaction tokens need to be enriched with the global motion information from the video tokens (green blocks).}
    \label{fig:motivation}
\end{figure}

Humans can recognize the action of \textit{pouring something into something} regardless of the objects involved, whether it involves water into a glass or sauce into a saucepan. 
The reason humans can do this is because they can naturally disentangle the objects from their motion and context, a process known as compositionality.
In literature, action compositionality may refer to part-whole relationships where a complex action comprises basic atomic actions \cite{girdhar2019cater}. In this paper, we investigate the object-agnostic aspect of compositional action recognition.
To test a system's compositionality, authors in \cite{materzynska2020something} formulate compositional action recognition as the task of recognizing actions involving objects unseen during training.
Compositional action recognition requires understanding human interaction with objects disentangled from the background and appearance biases of the objects.

Hence, we build a system (as shown in \cref{fig:motivation}) to explicitly model human-object interactions from the spatial location features extracted from the respective humans and objects. 
Unlike object-centric representation, interaction-centric representation is better suited since interactions are less affected by object-centric inductive biases for compositional actions that are by definition \textit{object-agnostic}.
We then impart the interaction representation with global motion information to infuse vital context that can help differentiate between similar compositional actions leading to more accurate recognition. 

Compositional and fine-grained action recognition has been addressed in literature by modeling objects, humans, and their interactions~\cite{kim2021motion, zhang2022object, herzig2022object, yan2023progressive}. 
Compared to successful action-recognition models that capture the spatio-temporal changes of appearance in videos~\cite{carreira2017quo,feichtenhofer2019slowfast,bertasius2021space,patrick2021keeping}, a common thread among these methods is to build an object-centric video representation of the \textit{active} objects~\cite{zhang2022object,herzig2022object} i.e., objects involved in interactions.
Object features are also used in \cite{kim2021motion} to guide the attention of appearance features.
All these approaches model the changes in the appearance and spatial positions of active objects and the learned object representations are then fused with the video-level representation. 
These methods are able to capture object-centric inductive biases induced by the change in the appearance and spatial locations of the objects for effective action recognition.
However, for generalizing action understanding across to unseen objects in compositional action recognition, learning object-agnostic \textit{interaction representations} is more important. 
Therefore, we propose to encode interactions involving hands and objects from the appearance and spatial location features extracted from the respective humans and objects. 
Unlike object-centric representation, interaction-centric representation is better suited as interactions are less affected by object-centric inductive biases for compositional actions which are object-agnostic.

Modelling interactions between humans, objects and stuff involves capturing the dynamics of their fine-grained relationships~\cite{ou2022object}. Prior work on human-object interactions have either looked at modelling human-object relationships~\cite{ji2021detecting} using relationship labels or detecting/forecasting~\cite{liu2020forecasting,li2021weakly,liu2022joint} a specific interaction region. Instead in this work, we focus on modelling all possible human-object and object-object and human-object-stuff interactions to identify the important interactions.
In~\cite{roy2024interaction}, an interaction modelling framework is proposed for action anticipation. 
While ~\cite{roy2024interaction} only considers appearance features, we explicitly encode the frame-wise coordinates of humans and objects vital for object-agnostic compositional action recognition.
Furthermore, we refine the interaction-centric representation with global motion which contains con-
textual cues to differentiate similar actions such as \textit{rolling down slanted surface} with moving something
down or lifting up one end of something, then letting it drop down.

To model interactions, existing approaches use either intra-frame (\cite{kim2021motion}) or inter-frame interactions (\cite{roy2024interaction}).
Frame-wise hand-object interactions capture fundamental building blocks of hand-object interactions (like grasping, pointing, pushing) even if those specific movements haven't been seen in combination with the test object during training.
However, frame-wise interactions can struggles to understand longer-term relationships or compositional actions that require a sequence of interactions.
On the other hand, long-range hand-object interaction may not be able to identify sub-actions but it can disambiguate actions that might look similar in a single frame.
For example, hand-object position in one frame could be part of lifting an object in one case and putting it down in another, depending on the context.
Therefore, we propose a fused frame-trajectory interaction encoder that models both the fine-grained interactions between hand and objects in each frame, and long-range (entire video) interactions between hand and object trajectories over the entire video. 
We fuse the information from both these interactions into spatio-temporal interaction tokens (\sti).
We also propose a global motion infusion stage that refine the \sti iteratively in multiple layers of a video transformer based on \cite{patrick2021keeping}.
Inspired by \cite{kim2021motion}, our proposed global motion infusion can explore higher-level correlations
between interaction tokens and global motion.
We show that both the interaction encoding stage and global fusion stage are complimentary and lead to massive gains in compositional action recognition on Something Else dataset.

\textbf{Contributions}: We design an object agnostic human-object interaction encoder for compositional action recognition that operates on human and object coordinate features. The object-agnostic interaction representation infused with the global motion information sets a new state-of-the-art on the Something-Else dataset.

\section{Related Work}
\label{sec:related-work}

\textbf{Interaction modelling:}
There is a wealth of literature for modelling human and object interactions in images~\cite{chao2018learning, gao2018ican,li2019transferable,gao2020drg, zhang2021mining, chen2021reformulating, kim2020uniondet, kim2021hotr,liao2020ppdm}. However, interaction modelling for videos has not received enough attention ~\cite{liu2020forecasting,li2021weakly,liu2022joint,ji2021detecting}. When compared to modelling human-object interactions in images, focussing on the active objects among all the background objects in videos is hard. Models should also be able to capture the temporal dimension of the action in case of \textit{pushing/pulling something from left to right/right to left, moving something up/down}. The interaction modelling approaches developed for 2D static images cannot be extended naively to 3D videos with motion blur. Given this context, of particular interest to us are  ~\cite{li2021weakly, ji2021detecting}. ~\cite{li2021weakly} develop a contrastive loss to detect human-object interactions in a weakly supervised manner (without bounding box annotations) given verb-object pairs extracted from video captions as label. ~\cite{ji2021detecting} build a network of densely connected Transformers for detecting human-object relationship triplets in videos. Object distributions are typically long-tailed and common human-object interactions occur more often than others. Our focus is on modelling interactions for compositional actions on unseen objects between train and test time. 

\textbf{Compositional Action Recognition:}
Conventional action recognition involves learning coarse video-level appearance features.
In order to mitigate the effect of overfitting on the video-level appearance features, a wide variety of object-centric methods ~\cite{yan2020interactive, sun2018actor, wang2018videos, xu2020spatio, baradel2018object, ji2020action, materzynska2020something, li2021weakly, girdhar2019video, zhang2019structured, wu2019long} have been developed. ~\cite{chen2019graph, wang2018non, herzig2019spatio, ji2020action, wang2018videos, arnab2021unified} model videos as space-time graphs. In object-centric methods, the localized appearance features of the humans and objects are often extracted using RoI-pooling~\cite{he2017mask, faster2015towards} of bounding boxes from an object detector. These features are then temporally aggregated using recurrent neural networks~\cite{yan2020interactive, baradel2018object, ma2018attend}, graph convolutional networks ~\cite{wang2018videos}, attention mechanisms ~\cite{ma2018attend, wu2019long, perez2020knowing, ji2020action}. The learned human and object representations are then fused~\cite{sun2018actor, wang2018videos, materzynska2020something} with the global video level features. Different from the object-centric methods, our method is interaction-centric. We first build an interaction representation. The interaction representation fused with context information is then be used for compositional action recognition~\cite{materzynska2020something, girdhar2019cater, ji2020action}.
In \cite{yan2023progressive}, features that encode only the position of the hand and object are learned separately appearance features of hand and objects. 
Both these features are used to reason about the role of each object in the action and how the objects and hand interact during an action.
The reasoning allows prediction of unseen compositional actions with novel hand object combinations.
In our setting, the essence of compositionality lies in modelling the relationships between humans and objects so as to recognize actions with unseen combinations of verbs and nouns.

\textbf{Video Transformers:}
Video Transformers extend vision Transformer backbones to time domain through spatial and temporal attention mechanisms. Timesformer~\cite{bertasius2021space}, ViViT~\cite{arnab2021vivit}, MViT~\cite{fan2021multiscale}, MeMViT~\cite{wu2022memvit}, Motionformer~\cite{patrick2021keeping} and ORViT~\cite{herzig2022object} have emerged as powerful baselines for modelling videos.
Closest to our line of work, ORViT~\cite{herzig2022object} augments Motionformer~\cite{patrick2021keeping} video representation with object appearance features. Our approach refines the learned interaction representation with information from the video representation. Recently, ~\cite{zhang2022object} develop an object-centric video Transformer that uses cross-attention between summarized object representations and the video representation to learn an object-centric video representation. We take an interaction-centric approach where the cross attention is performed on spatio-temporal interaction representation and video context representation.

\section{Modelling Interactions Involving Hands and Objects}
\begin{figure}
    \centering
    \includegraphics[width=\linewidth]{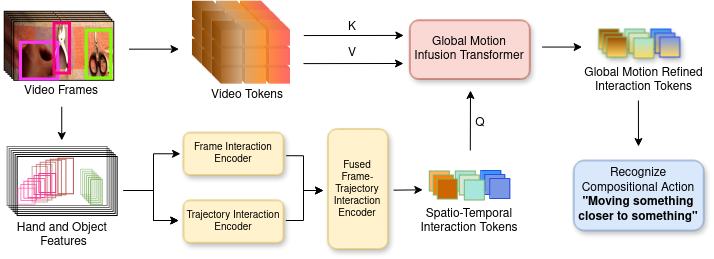}
    \caption{We first extract the hand and object features (coordinate and appearance) from 2D frames. The Frame Interaction Encoder encodes fine-grained hand-object interactions while the Trajectory Interaction Encoder encodes long-range interactions between hand and object trajectories. 
    The output of these two interactions is fused to form Spatio-Temporal-Interaction \sti. \sti\ are then refined with global motion information from video tokens. We use the global motion infused \sti\ for compositional action recognition.}
    \label{fig:overview}
\end{figure}

First, we present the features required for interaction modeling in \cref{sec:prelims}. Then in \cref{sec:overview}, we present the overview of our interaction modelling approach for compositional action recognition.
In \cref{sec:sti_enc}, we detail the interaction encoder that captures the fine-grained and into spatio-temporal interaction tokens (\sti).
In \cref{sec:cattn}, we describe a visual transformer model that refines the \sti with global motion.
The overview of our transformer based interaction modeling approach, for compositional action recognition is shown in \cref{fig:overview}.

\subsection{Hand and Object Features}
\label{sec:prelims}
As each video contains at least one human action, we assume there is at least one hand and $M$ objects in each frame.
We denote human and object representations by $\mathbf{h}_{t} \in \mathbb{R}^d$ and $\mathbf{o}_{i,t} \in \mathbb{R}^d$, $i \in M$ respectively for each frame at index $t$ where $t = \{1, \cdots, T \}$.
In particular, we obtain these d-dimensional  representations by concatenating appearance and coordinate (spatial location) features. 
The appearance features are extracted using the bounding box coordinates via a RoIAlign~\cite{he2017mask} layer followed by average-pooling similar to~\cite{herzig2022object}. 
The coordinate features are obtained by projecting the center coordinates, width, and height of their respective bounding boxes to a ${d}-$dimensional embedding. 
We then apply a multi-layer perceptron (MLP) on the concatenated appearance and coordinate vectors to obtain the final vector representation.
In case a hand/object is missing in a frame, we use the bounding box of the hand/object from the previous frame as coordinates and its content as appearance features.
As almost all the actions in Something Else \cite{materzynska2020something} are performed by one hand, we consider only one hand per frame to maintain uniformity across actions.

\subsection{Overview of hand-object interaction modeling}
\label{sec:overview}
We build on the human and object representations defined in~\cref{sec:prelims} to model their interactions. Human-object and object-object interactions are important for compositional action recognition. We encapsulate interactions using an encoder function. This generic encoder function takes human and object features (coordinates, appearance or both as defined in \cref{sec:prelims}) as inputs and outputs either joint representations,  disjoint interaction representations.
We call this encoder as fused frame-trajectory interaction encoder (described in \cref{sec:sti_enc}).
Let us denote the output of our fused frame-trajectory interaction encoder $\psi$ in Equation~\eqref{eq:encoder} by $\mathbf{z}$.
\begin{equation}
\mathbf{z} = \psi(\mathbf{h}, \mathbf{o}_{i} )~\forall~i
\label{eq:encoder}
\end{equation}
Overall, $\mathbf{z}$ captures how the humans and objects affect and interact with each other and embed the interactions into the refined representations while encapsulating relevant information about the compositional actions.
\section{Fused Frame-Trajectory Interaction Encoder}
\label{sec:sti_enc}
\begin{figure}
    \centering
    \includegraphics[width=0.80\linewidth]{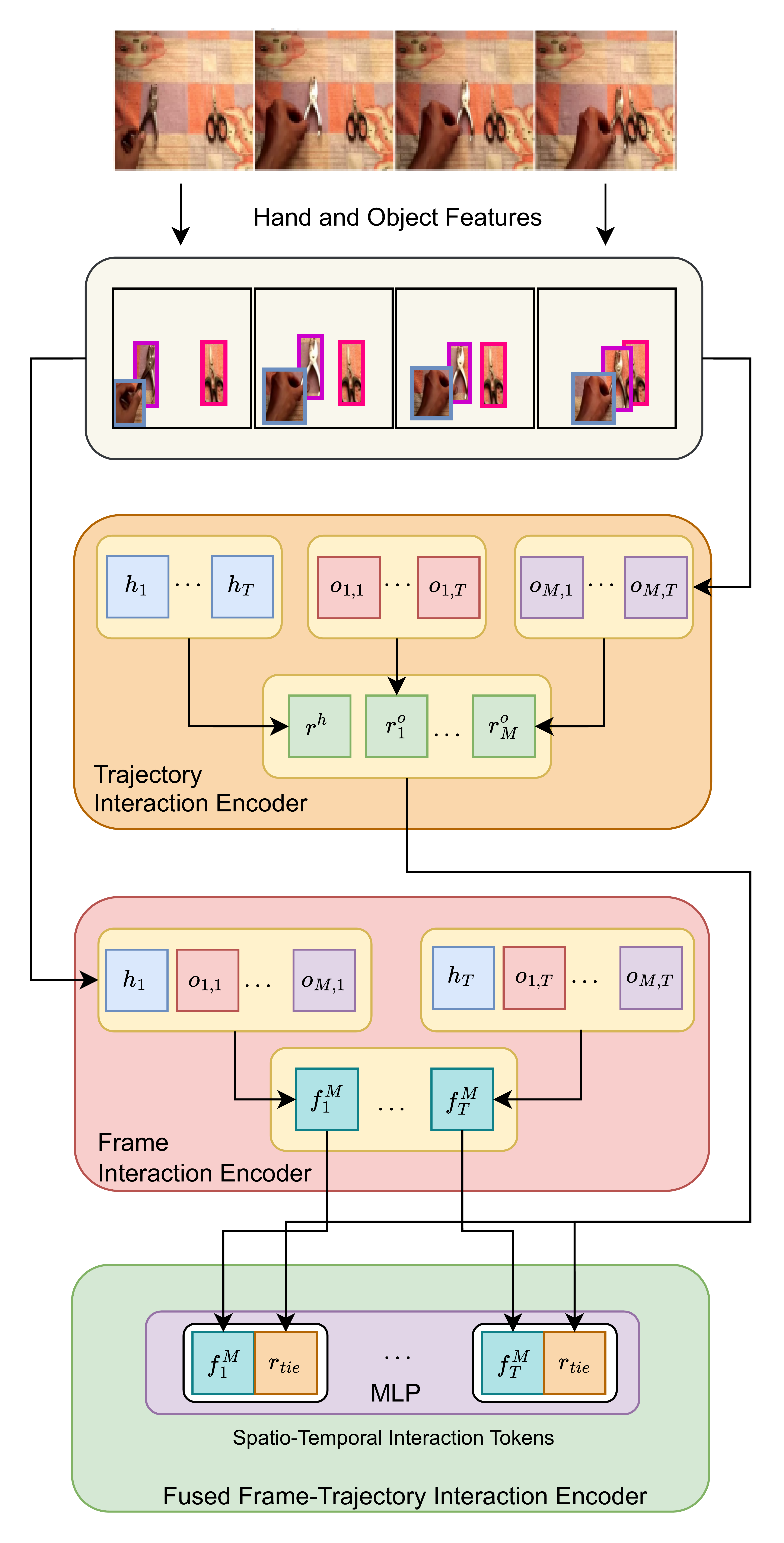}
    \caption{Block diagram of the Fused Frame-Trajectory Interaction Encoder. Fine-grained interactions between the hand and the object features in every frame are modelled via the Frame Interaction Encoder. Long-range interactions between the hand and the object feature trajectories across the video are modelled using the Trajectory Interaction Encoder. The outputs of the two parallel encoders are fused using an MLP to obtain the \sti.}
    \label{fig:ffte}
\end{figure}

Given a video, we want to encapsulate the interactions disentangled from the background into spatio-temporal interaction tokens or \sti\ denoted by $\mathbf{z}$ above. 
For this purpose, we design the Fused Frame-Trajectory Encoder to fuse the information from fine-grained interactions in a frame and long-range interactions between hand and objects over the video.
For fine-grained interactions, we model the interactions between hand-object bounding boxes at every frame.
For long-range interactions, we model the interactions between hand trajectories and object trajectories across the video.
These two interactions are computed in two parallel networks -- Frame Interaction Encoder and Trajectory Interaction Encoder. The output of these networks is fused to produce the \sti.
The block diagram of the Fused Frame-Trajectory Interaction Encoder is shown in \cref{fig:ffte}.

\subsection{Frame Interaction Encoder (FIE)}
To capture the hand-object and object-object interactions at the frame level, we take the sequence of hand and object representations $\{ \mathbf{h}_t, \mathbf{o}_{1,t}, \cdots, \mathbf{o}_{M,t} \}$ as input.
FIE obtains a frame representation $\mathbf{f}_t$ that summarizes the intra-frame interactions between hand and objects.
\begin{equation}
    \mathbf{f}_t = \text{FIE}(\mathbf{h}_t, \mathbf{o}_{1,t}, \cdots, \mathbf{o}_{M,t})
    \label{eq:local_motion_sf}
\end{equation}
We realize FIE with both BLSTM and Transformer as they are both sequence summarization models and have their own advantages.
While BLSTM captures the interactions between the hand and objects in the frame sequentially while Transformer captures all the interactions together through self-attention. 
We compare the effect of using BLSTM and Transformer in every stage of our encoding in \cref{sec:ablation}.

\textit{BLTSM}$_F$: takes the sequence of  hand and object features at every frame as tokens and returns a sequence of hidden states as outputs. 
We consider the final hidden state $\mathbf{f}_t^{M}$ as the frame summary for the intra-frame interactions between the hand and the objects.
\begin{align}
     \mathbf{f}_t^1, \cdots, \mathbf{f}_t^{M}  &= \text{BLSTM}( \mathbf{h}_t, \mathbf{o}_{1,t}, \cdots, \mathbf{o}_{M,t} ) \nonumber \\
    \mathbf{f}_t &=  \mathbf{f}_t^{M}
\end{align}

\textit{Transformer}$_F$: takes the hand and object features at every frame as tokens and models the interactions using self-attention ~\cite{vaswani2017attention}. 
A learnable positional encoding $\mathbf{p} = \mathbb{R}^{M+1 \times d}$ is added to the hand and object features maintain the order that the hand feature is considered first followed by object features. 
We add a learnable token $\mathbf{l}_t$ that captures the frame-wise summary of the hand-object interactions. 
\begin{align}
      \hat{\mathbf{l}}_t, \hat{\mathbf{h}}_t, \hat{\mathbf{o}}_{1,t}, \cdots, \hat{\mathbf{o}}_{M,t}  &= \text{Transformer}(\mathbf{l}_t, \mathbf{h}_t, \mathbf{o}_{1,t}, \cdots, \mathbf{o}_{M,t} )\nonumber\\
     \mathbf{f}_t &= \hat{\mathbf{l}}_t 
    \label{eq:lstm_sf}
\end{align}
The transformer self-attention as per ~\cite{vaswani2017attention} is computed as
\begin{equation}
 \hat{\mathbf{x}} = \sum \mathbf{v} \frac{\text{exp}\left< \mathbf{q}, \mathbf{k}\right>}{\sum {\text{exp}\left< \mathbf{q}, \mathbf{k} \right>}} 
\label{eq:sa-attn}
\end{equation}
where $\mathbf{x} =\{\mathbf{l}_t, \mathbf{h}_t, \mathbf{o}_{1,t}, \cdots, \mathbf{o}_{M,t}$\}, $\mathbf{q} = \mathbf{x} \mathbf{W}_{q}, \mathbf{k} =  \mathbf{x} \mathbf{W}_{k}, \mathbf{v} = \mathbf{x} \mathbf{W}_{v}$, and $\mathbf{W}_{q}, \mathbf{W}_{k}, \mathbf{W}_{v} \in \mathbb{R}^d \times \mathbb{R}^d$.
In the subsequent sections, every Transformer() follows the formulation in \cref{eq:sa-attn} unless described otherwise.

\subsection{Trajectory Interaction Encoder (TIE)}
To capture long-range interactions between hands and objects over the entire video, we model the interaction between the trajectories of hands and objects. 
The TIE model takes a sequence of hand features over time $ \mathbf{h}_1, \cdots \mathbf{h}_T $ and object features over time $ \mathbf{o}_{m,1}, \cdots, \mathbf{o}_{m,T} ; m=\{1, \cdots, M\}$ as input.
The TIE model returns the summarized long-range interaction of hand and object trajectories in the tokens $\mathbf{r}^{h}, \mathbf{r}^{o}_{1}, \cdots, \mathbf{r}^{o}_{M}$.
\begin{equation}
\begin{split}
    \mathbf{r}^{h}, \mathbf{r}^{o}_{1}, \cdots, \mathbf{r}^{o}_{M} &= \text{TIE}
    \left(
    \begin{aligned}
    &  \mathbf{h}_1, \cdots, \mathbf{h}_{T} , \\
    &  \mathbf{o}_{1,1}, \cdots, \mathbf{o}_{1,T} , \\
    & \cdots, \\
    &  \mathbf{o}_{M,1}, \cdots, \mathbf{o}_{M,T}  \\
    \end{aligned} 
    \right )
\end{split}
\end{equation}
Again, as with frame interaction encoder, we realize the TIE model with both BLSTM and Transformer.

 \textit{BLSTM}$_T$: Each hand and object trajectory is model separately using BLSTMs and we obtain trajectory summaries $\hat{\mathbf{h}}_T$ and $\hat{\mathbf{o}}_{m,T}$, respectively.
\begin{align}
     \hat{\mathbf{h}}_1, \cdots, \hat{\mathbf{h}}_T  = \text{BLSTM}( \mathbf{h}_1, \cdots, \mathbf{h}_{T} ) \nonumber \\
     \hat{\mathbf{o}}_{m,1}, \cdots, \hat{\mathbf{o}}_{m,T}  = \text{BLSTM}( \mathbf{o}_{m,1}, \cdots, \mathbf{o}_{m,T} ) \nonumber
\end{align}
Another BLSTM network then takes the summarized hand and object trajectory representations to obtain long-range interaction tokens as follows:
\begin{align}
     \mathbf{r}^{h}, \mathbf{r}^{o}_{1}, \cdots, \mathbf{r}^{o}_{M} 
    = \text{BLSTM}( \hat{\mathbf{h}}_T, \hat{\mathbf{o}}_{1,T}, \cdots, \hat{\mathbf{o}}_{M,T}).
    \label{eq:lstm-ts}
\end{align}

\textit{Transformer}$_T$: models the temporal interaction separately between hand features over time or object features over time. 
The learnable tokens $\mathbf{l}_h$ and $\mathbf{l}_{o,m}$ learn the summarized trajectory interaction for hand and object $m$, respectively.
\begin{align}
     &\hat{\mathbf{l}}_h, \hat{\mathbf{h}}_1, \cdots, \hat{\mathbf{h}}_T  = \text{Transformer}(\mathbf{z}, \mathbf{h}_1, \cdots, \mathbf{h}_{T}  ) \nonumber \\
     & \hat{\mathbf{l}}_{o,m}, \hat{\mathbf{o}}_{m,1}, \cdots, \hat{\mathbf{o}}_{m,T}  \nonumber \\ 
     &= \text{Transformer}( \mathbf{l}_{o,m}, \mathbf{o}_{m, 1}, \cdots, \mathbf{o}_{m,T} ) 
\end{align}
With the summarized hand and object trajectory representations, we use another lightweight transformer to compute the long-range interaction tokens as follows:
\begin{equation}
\mathbf{r}^{h}, \mathbf{r}^{o}_{1}, \cdots, \mathbf{r}^{o}_{M}  = \text{Transformer}(\hat{\mathbf{l}}_h, \hat{\mathbf{l}}_{o,1}, \cdots, \hat{\mathbf{l}}_{o,M}).
\label{eq:trans_t}
\end{equation}

\subsection{Fusing Frame and Trajectory Interactions}
The frame-level hand-object interactions from FIE provide $T$ frame-level summary tokens while the trajectory-level interactions from TIE provide $M+1$ summary tokens.
The trajectory level interaction summary is video-level information that needs to be added to every frame-level interaction token. 
Therefore, we choose to summarize $M+1$ TIE summary tokens using a BLSTM or Transformer to a single token $\mathbf{r}_{tie}$ that is added to the frame-level interaction tokens. 

\textit{BLSTM}$_{FT}:$ In case of BLSTM, $\mathbf{r}^{h}, \mathbf{r}^{o}_{1}, \cdots, \mathbf{r}^{o}_{M} $ is the input and the final hidden state is considered as the summary $\mathbf{r}_{tie}$. 

\textit{Transformer}$_{FT}:$For Transformer, we append a learnable token to $\mathbf{r}^{h}, \mathbf{r}^{o}_{1}, \cdots, \mathbf{r}^{o}_{M} $ and take the refined output of the learned token as the summary $\mathbf{r}_{tie}$.

Next, we concatenate a trajectory-level interaction summary $\mathbf{r}_{tie}$ to the frame-level interaction summaries.
\begin{equation}
\underbrace{\mathbf{z}_1, \cdots, \mathbf{z}_T }_{\mathbf{z}} = \text{MLP}({[\mathbf{f}_1 ; \mathbf{r}_{tie}]}, \cdots, [ \mathbf{f}_T;\mathbf{r}_{tie}])
\end{equation}
where [;] is the concatenation operator and the MLP is used to reduce the concatenated feature dimension from $\mathbb{R}^{2d}$ to $\mathbb{R}^{d}$.
\section{Infusing Global Motion Into Interactions}
\label{sec:cattn}

The interaction-refined \sti\ $\mathbf{z}$ capture the fine-grained and long-range interactions as correlations between the human and the object features. 
These \sti\ can be further refined with global motion which contains contextual cues to differentiate similar actions as illustrated in ~\cref{fig:motivation}. 
Therefore, we infuse the context-disentangled \sti\ with global motion.

Let $X \in \mathbb{R}^{THW\times 3}$ be the set of all frames in a given input video. 
$H$, $W$ and $T$ are the height, width, and number of frames of $X$, respectively. 
We adopt the token extraction framework from video Transformers~\cite{arnab2021vivit,patrick2021keeping,bertasius2021space} to construct $X' \in \mathbb{R}^{T'H'W' \times d}$ ``video tokens''. 
$T'H'W'$ is the number of all video tokens which are cuboidal embedding of $X$ projected to $d$ dimensions and each token is denoted by $\mathbf{x}_{st} \in \mathbb{R}^d$ where $s$ is the spatial position and $t$ is the temporal frame position.
In order to update the \sti\ ($\mathbf{z}$) with the information contained in video tokens $\mathbf{X'}$, we exploit the attention mechanism to pool information along motion paths from Motionformer~\cite{patrick2021keeping} in \textit{Global Motion Infusion Transformer}.

\sti\ $\mathbf{z}$ act as queries while $\mathbf{X'}$ 
video tokens acting as keys and values.
For each \textit{STI token}, we perform attention along the probabilistic path --trajectory-- of the $\mathbf{z_t}$ tokens between frames. The set of trajectory tokens  denoted by $\hat{\mathbf{z}}_{tt'} \in R^{d}$ at different times $t'$ represents the pooled information weighted by the trajectory probability. The attention is applied spatially and independently for each \textit{STI token} in each frame.
Equation~\eqref{eq:traj-attn-s} gives the pooling operation that implicitly seeks the location of the interaction trajectory at time $t'$ by
comparing the interaction query $\mathbf{q}_t = \mathbf{z}_t \textbf{W}_q$ to the global motion keys $\mathbf{k}_{st'} = \textbf{x}_{st'} \textbf{W}_k$ while the values are given by $\mathbf{v}_{st'} = \textbf{x}_{st'} \textbf{W}_v$.

\begin{figure}
    \centering
    \includegraphics[width=\linewidth]{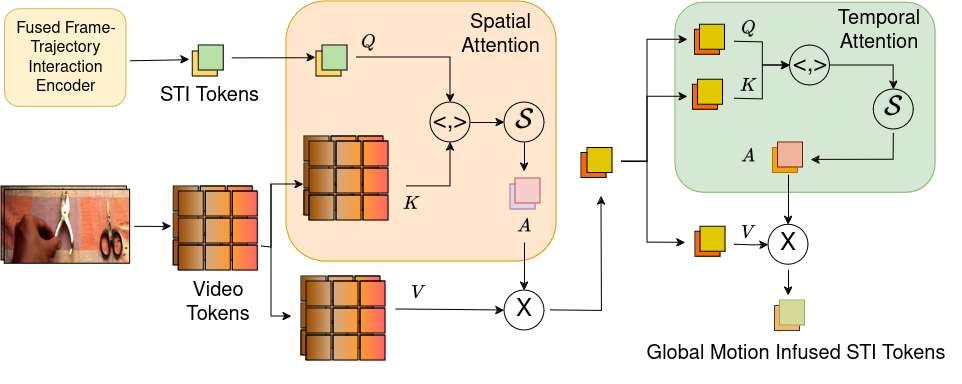}
    \caption{\sti\ are infused with global motion via the Global Motion Infusion Transformer. By matching the \textit{STI token} queries with the video token keys, the spatial attention operation first computes the best location for the \textit{STI token} trajectories. Next, the temporal attention operation performs pooling of the interaction trajectories across time to accumulate the temporal information in \sti. The global motion infused \sti\ are then used for compositional action recognition. $\mathcal{S}$ denotes the softmax function, $<,>$ denotes the inner-product operator and $X$ the denotes weighted sum operator.}
    \label{fig:context-infusion}
\end{figure}

\begin{equation}
\hat{\mathbf{z}}_{tt'} = \sum_{s} \mathbf{v}_{st'} \frac{\text{exp}\left< \mathbf{q}_t, \mathbf{k}_{st'}\right>}{\sum_{\bar{s}} {\text{exp}\left< \mathbf{q}_t, \mathbf{k}_{\bar{s}t'} \right>}} 
    \label{eq:traj-attn-s}
\end{equation}

Once the trajectories are computed, we pool them across time to reason about the intra-frame connections across the interaction regions. This is achieved by projecting the trajectory tokens to a new set of queries, keys and values $\hat{\mathbf{q}_t} = \hat{\mathbf{z}}_{tt'} \textbf{W}_q$, $\hat{\mathbf{k}}_{tt'} = \hat{\mathbf{z}}_{tt'} \textbf{W}_k$ and $\hat{\mathbf{v}}_{tt'} = \hat{\mathbf{z}}_{tt'} \textbf{W}_v$. This new query $\hat{\mathbf{q}_t}$ contains information from the entire interaction trajectory across video frames. Equation~\eqref{eq:traj-attn-t} gives the pooling operation across the new time (trajectory) dimension by applying 1D attention using $\hat{\mathbf{q}_t}$.

\begin{equation}
\widehat{\mathbf{z}}_{t} = \sum_{t'} \hat{\mathbf{v}_{tt'}} \frac{\text{exp}\left< \hat{\mathbf{q}}_t, \hat{\mathbf{k}}_{tt'}\right>}{\sum_{\bar{t}} {\text{exp}\left< \hat{\mathbf{q}}_t, \hat{\mathbf{k}}_{t\bar{t}} \right>}} 
    \label{eq:traj-attn-t}
\end{equation}
In order to predict the compositional action, the global motion infusion is coupled with the trajectory attention block of Motionformer~\cite{patrick2021keeping} to iteratively refine the \sti. 
We pass the final hidden-state of $\mathbf{z}$ from the output of the local interaction encoder described in \cref{sec:sti_enc} to the global motion infusion Transformer. The modified class token $\mathbf{z}'_{cls}$ obtained after 12 layers of cross attention is passed through a linear layer to get the predicted action. We train our models with cross-entropy loss as the objective function.
\section{Experimental Results}
\label{sec:experiments}
In this section, we present the performance on the compositional action dataset Something-Else ~\cite{materzynska2020something}. The compositional split of the Something-Else dataset is designed for generalization across objects such that the objects in the training set (54,919 samples) are different from the objects in the validation set (54,876 samples). The dataset also contains the ground-truth bounding box annotations for all the humans and objects.

\subsection{Implementation Details}
We uniformly sample clips of 224x224 resolution and length of 16 frames. We follow ~\cite{herzig2022object} to choose the number of objects by taking the maximum number of objects per video across all videos in the training set. We set the number of hand to 1 and the maximum number of objects to 3. If fewer than 3 objects or 1 hand is found in a frame, the respective bounding boxes and feature vectors are padded with the features from the previous frame. If there are more than one hand in a frame, we randomly choose one. 

\textbf{Interaction Encoder}
The human and object inputs are coordinate-features of dimension size 256. For the experiments where appearance features are also used, the 512 dimension appearance features are concatenated with the co-ordinate features to obtain a 768 dimensional vector.    

The $\text{BLSTM}_{FT}$ model consists of a single layer of spatial BLSTM followed by another single layer of temporal BLSTM. The input to the temporal BLSTM is summarized hidden-state vectors from the spatial BLSTM of size 768. The size of the hidden-state dimension of both the spatial BLSTM and the temporal BLSTM is set to 768. 
Each transformer model $\text{Transformer}_{FT}$ is formed of 6 transformer encoder layers each with 12 attention heads and we use 768 as the embedding dimension size. 
We trained our transformer and BLSTM models with a batch size of 128 and a learning rate of $1e^{-3}$ for 50 epochs.

\textbf{Infusing Global Motion into Interactions:}
The \sti\ are used as queries and  the video tokens as keys and values. After each time the \sti\ are refined, the refined \sti\ are used as queries on the video tokens. This is followed by Motionformer's trajectory attention on the video tokens.
We apply 12 layers of the Global Motion Infusion Transformer with 12 attention heads and a temporal resolution of 8 to obtain the global motion infused \sti~\cref{sec:cattn}. The video tokens are constructed from 2x16x16 size cuboidal patches from the 224x224 resolution clips of length 16 frames. Learning rate of the global-motion-infusion module is set to $5e^{-5}$ and we trained our models using a batch size of 20. 

AdamW optimizer is used across all models and we conduct our experiments on 2 NVidia-RTX A6000 GPUs with 48 GB memory each.

\subsection{Ablation on Interaction Encoder}
\label{sec:ablation}
We study the effect of changes in the local positions of human-object interactions for compositional action recognition by fixing the coordinate embeddings and exploring the architecture design for the BLSTM and Transformer encoders in \cref{tab:lstm-ablations}. We compare the performance of the encoders on the Something-Else dataset.

First, we consider the Frame-Interaction $\text{BLSTM}_{F}$(C) with coordinate features for humans and objects as described in \cref{sec:sti_enc}. The interactions are first summarized using a spatial BLSTM at the frame-level and then the interaction trajectories are learnt across frames using a temporal BLSTM.

In the $\text{BLSTM}_{T}$(C) model, the individual human and object feature trajectories are first summarized across time using a temporal BLSTM and subsequently the interactions between the summarized trajectories are modelled via a spatial BLSTM.

In our experiments, $\text{BLSTM}_{F}$(C) performs slightly better than the late fusion approach in $\text{BLSTM}_{T}$(C). 
Next, we combine both the Frame-Interaction BLSTM and the Trajectory-Interaction BLSTM in Fused-Frame-Trajectory Interaction $\text{BLSTM}_{FT}$(C) as described earlier in \cref{sec:sti_enc}. This model outperforms the individual $\text{BLSTM}_{F}$(C) and $\text{BLSTM}_{T}$(C) models achieving 53.8\% top-1 accuracy. This can be attributed to the effect of the individual trajectories of the humans and the objects acting as a regularizer to the Frame-Interaction BLSTM. Similarly we observe that Fused-Frame-Trajectory Interaction $\text{Transformer}_{FT}$ Interaction model captures the spatio-temporal interactions effectively than the $\text{Transformer}_{F}$ and $\text{Transformer}_{T}$ models.

\begin{table}[!htbp]
\centering
\resizebox{0.75\linewidth}{!}{
\begin{tabular}{lcc} 
\hline
Method & Top-1(\%) & Top-5(\%) \\ \hline
$\text{BLSTM}_{F}$(C) & 49.4 & 77.5 \\
$\text{BLSTM}_{T}$(C) & 46.6 & 75.7 \\ \hline
$\text{Transformer}_{F}$(C) & 52.2 & 79.6 \\
$\text{Transformer}_{T}$(C) & 49.5 & 77.9  \\ \hline
$\text{BLSTM}_{FT}$(C) & \textbf{53.8} & \textbf{79.3} \\  
$\text{BLSTM}_{FT}$(C, A) & 53.3 & 79.1 \\ 
\hline

$\text{Transformer}_{FT}$(C) & 53.8 & 81.0 \\  
$\text{Transformer}_{FT}$(C, A) & \textbf{54.6} & \textbf{81.7} \\ \hline
\end{tabular}}
\caption{Comparison of Interaction Encoder with BLSTM and Transformer implementation on Something-Else dataset. C: Coordinate Embeddings, A: Appearance. Fused-Frame-Trajectory Interaction Encoder with Transformer achieves the best top-1 accuracy compared to its BLSTM counterpart.}
\label{tab:lstm-ablations}
\end{table}
 
\subsection{Effect of Infusing Global Motion}

We enrich the \sti\ obtained from the spatio-temporal interaction encoder using our proposed Global Motion Infusion. \cref{tab:context-infusion} demonstrates the effectiveness of our method in the  compositional action recognition task on the Something-Else dataset.
We compare the performance of the spatio-temporal interaction encoder on the best-performing set of $\text{BLSTM}_{FT}$ and $\text{Transformer}_{FT}$ encoders that captures the human-object interactions.
Our method achieves state-of-the-art results with \sti\ from the $\text{BLSTM}_{FT}$(C,A) encoder outperforming the $\text{Transformer}_{FT}$(C,A) model by +3.2\% on the Top-1 accuracy metric. 
It is interesting to note that after global motion infusion, the $\text{Transformer}_{FT}$ encoder performance is outperformed by the $\text{BLSTM}_{FT}$ encoder.

\begin{table}[!htbp]
\centering
\resizebox{0.8\linewidth}{!}{
\begin{tabular}{lccc} 
\hline
Method & Global Motion & Top-1(\%) & Top-5(\%) \\ \hline
$\text{Transformer}_{FT}$(C,A) & $\times$ & 54.6 & 81.7 \\ 
$\text{Transformer}_{FT}$(C,A) & \checkmark & 81.6 & 96.3 \\ \hline 
$\text{BLSTM}_{FT}$(C,A) & $\times$ & 53.3 & 79.1  \\
$\text{BLSTM}_{FT}$(C,A) & \checkmark & \textbf{84.8} & \textbf{97.2}  \\ \hline

\end{tabular}
}
\caption{Performance of global motion infusion into \sti\ (Something Else dataset). Global motion infusion substantially enriches the performance of the \sti\ obtained from the Fused-Frame-Trajectory Interaction Encoder models with state-of-the-art top-1 and top-5 accuracy.}
\label{tab:context-infusion}
\end{table}

\subsection{Comparison with State-of-the-Art}

We evaluate the performance of our global motion infused \sti\ against other methods for compositional action recognition in Table~\ref{table:sota}. 
On the Something-Else dataset, our method - our model with coordinate features ($\text{BLSTM}_{FT}$(C) + Global Motion) performs significantly better than all previous models with an 9.3\% absolute improvement over \cite{huang2023semantic} which models object motion and their label representations extracted from text. Our model ($\text{BLSTM}_{FT}$(C) + Global Motion) achieves 10.0\% absolute improvement over \cite{zhang2022object} which combines object summary vectors based on tracked objects with Motionformer~\cite{zhang2019structured}.
ORViT~\cite{herzig2022object} does not consider humans and objects differently while our approach considers humans separately from objects during interaction modeling. 
This choice yields an absolute improvement of 13.9\% for our approach over ORViT on Something-Else.  
Other approaches such as STIN+I3D~\cite{materzynska2020something} combine appearance features and object identity embeddings in addition to the coordinate features. 
Our global motion infusion transformer with coordinate features alone fares much better (83.6\% vs. 54.6\%). 
We attribute the strong compositional action recognition performance to the explicit formulation of the spatio-temporal interaction trajectories as query tokens for Motionformer~\cite{patrick2021keeping}.
By explicitly encoding the refined coordinates and appearance features as queries, the trajectory attention mechanism in Motionformer gives less attention to object appearance. 
Finally, our model with combination of both coordinate and appearance features ($\text{BLSTM}_{FT}$(C,A) + Global Motion) performs slightly better than $\text{BLSTM}_{FT}$(C) + Global Motion model as the combined representation is able to query the global motion effectively to disambiguate similar actions. 
\begin{table}[!htbp]
\centering
\scriptsize
\begin{tabular}{lcc} 
\hline
Method & Top-1(\%) & Top-5(\%) \\ \hline
I3D-ResNet50~\cite{carreira2017quo} & 42.8 & 71.3 \\ 
SlowFast~\cite{feichtenhofer2019slowfast} & 45.2 & 73.4 \\ 
TimesFormer~\cite{bertasius2021space} & 44.2 & 76.8 \\ 
Motionformer~\cite{patrick2021keeping} & 60.2 & 85.8 \\ \hline
STIN+I3D~\cite{materzynska2020something} & 54.6 & 79.4 \\ 
STLT+R3D~\cite{radevski2021revisiting} & 67.1 & 90.4 \\ 
LRR~\cite{bhattacharyya2023look} & 62.0 & 86.3 \\ 
ORViT~\cite{herzig2022object} & 69.7 & 91.0 \\  
MF+STLT+OL~\cite{zhang2022object} & 73.6 & 93.5 \\ \hline
TEA~\cite{yan2023progressive} & 68.8 & 90.3 \\ 
DeFormer~\cite{huang2023semantic} & \underline{74.3} & \underline{93.7} \\ \hline 
$\text{BLSTM}_{FT}$(C) + Global Motion & 83.6 & 96.8  \\
$\text{BLSTM}_{FT}$(C,A) + Global Motion & \textbf{84.8} & \textbf{97.2} \\ \hline
\end{tabular}
\caption{Comparison of our method with state-of-the-art compositional action recognition baselines (Something-Else dataset). Our method outperforms recent  Transformer methods and even video reconstruction based methods \cite{yan2023progressive,huang2023semantic} by a significant margin. Best models are in \textbf{bold} while second best are \underline{underlined}.}
\label{table:sota}
\end{table}

\subsection{Qualitative Results}
\begin{figure}
\centering
\includegraphics[width=0.89\linewidth]{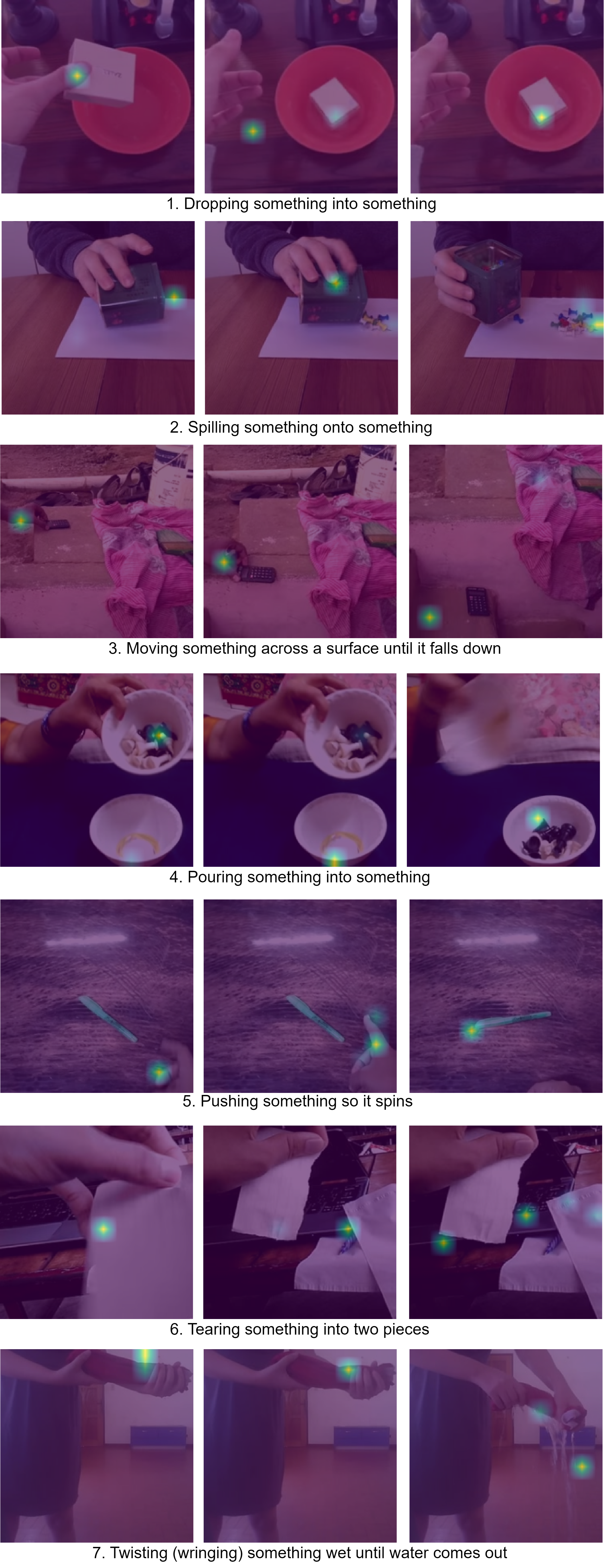}
\caption{Visualization of the attention of the final class token output on all the spatial tokens. In actions involving \textit{\{(3)Moving, (5)Pushing, (7)Twisting\} something}, our method focusses on the interaction regions. In \textit{\{(1)Dropping, (2)Spilling, (4)Pouring\} something into something} actions, our method pays high attention to both the objects involved in the interaction. In \textit{(6)Tearing something into two pieces}, our method attends to both the pieces after the object is torn.}
\label{fig:attn-map}
\end{figure}

We provide insight into the inner workings of our method in ~\cref{fig:attn-map}. We visualize the attention map of the final $\mathbf{z}'_{cls}$ output used to predict the compositional action on all spatial tokens.
Across all the examples, one can clearly see how human-object interactions affect the spatial attention maps.
In the examples (3)Moving something across a surface until it falls down, (5)Pushing something so it spins, and (7)Twisting something until water comes out, our method focusses on the interaction region between the human and the active object. This is despite the confounding background objects in the scene.
When there are two objects involved, our method focusses on both as can be seen in (1)Dropping something into something, (2)Spilling something onto something, (4)Pouring something into something. In (6)Tearing something into two pieces, our method attends both the pieces after the object is torn.

\section{Discussion and Conclusion}

Modelling interactions between humans and objects is a crucial step towards developing generalizable action recognition models.
In this work, we have presented an object agnostic human-object interaction method for compositional action recognition. Our approach takes human and object features as input and encodes them into a joint interaction representation. We have shown how the interaction representation can be enriched with global motion information that allows distinguishing similar actions. The geometric and spatial properties of objects (i.e., coordinates) are amazingly effective for compositional action recognition.

We demonstrate the effectiveness of our interaction encoder model on the compositional Something-Else dataset. We set a new state-of-the-art in compositional action recognition by outperforming recent important advancements by a significant margin indicating the significance of modelling spatial-temporal dynamics of interactions.

{\small
\bibliographystyle{ieee_fullname}
\bibliography{egbib}
}

\end{document}